\documentclass[conference,flushend]{iaria} 
\usepackage{caption}
\usepackage{multirow}
\usepackage{lscape}
\usepackage{booktabs, makecell} 
\usepackage{geometry}
\usepackage{pdflscape} 
\usepackage{graphicx}
\usepackage{xtab}
\usepackage{amsmath}
\usepackage{algorithm}
\usepackage{algpseudocode}
\usepackage{enumitem}
\usepackage{tikz}
\usepackage{subcaption}
\usepackage[caption=false,font=footnotesize]{subfig}
\usepackage{etoolbox} 

\newcommand\blfootnote[1]{%
  \begingroup
  \renewcommand\thefootnote{}\footnote{#1}%
  \addtocounter{footnote}{-1}%
  \endgroup
}

\usetikzlibrary{shapes.multipart, arrows.meta, positioning}

\title{From Bias to Balance: Fairness-Aware Paper Recommendation for Equitable Peer Review}

\author{
  \IEEEauthorblockN{%
    Uttamasha Anjally Oyshi, Susan Gauch\,}
  \IEEEauthorblockA{%
    Department of Electrical Engineering \& Computer Science\\
    University of Arkansas\\
    Fayetteville, USA\\
    e-mails: {\tt \{uoyshi, sgauch\}@uark.edu}
} }

\begin{document}
\maketitle

\blfootnote{\footnotesize This is an extended version of: U.\ A.\ Oyshi and S.\ Gauch,
``Fair Learning for Bias Mitigation and Quality Optimization in Paper Recommendation,''
17th Annual Conference on Information, Process, and Knowledge Management (eKNOW 2025),
Nice, France, May 18--22, 2025.}

\begin{abstract}

\noindent Despite frequent double-blind review, systemic biases related to author demographics still disadvantage underrepresented groups. We start from a simple hypothesis: if a post-review recommender is trained with an explicit fairness regularizer, it should increase inclusion \emph{without} degrading quality. To test this, we introduce \emph{Fair-PaperRec}, a Multi-Layer Perceptron (MLP) with a differentiable fairness loss over intersectional attributes (e.g., race, country) that re-ranks papers after double-blind review. We first probe the hypothesis on \emph{synthetic} datasets spanning high, moderate, and near-fair biases. Across multiple randomized runs, these controlled studies map where increasing the fairness weight strengthens macro/micro diversity while keeping utility approximately stable, demonstrating robustness and adaptability under varying disparity levels. We then carry the hypothesis into the \emph{original} setting—conference data from ACM Special Interest Group on Computer-Human Interaction (SIGCHI), Designing Interactive Systems (DIS), and Intelligent User Interfaces (IUI). In this real-world scenario, an appropriately tuned configuration of Fair-PaperRec achieves \emph{up to} a 42.03\% increase in underrepresented-group participation with at most a 3.16\% change in overall utility relative to the historical selection. Taken together, the synthetic-to-original progression shows that fairness regularization can act as both an equity mechanism and a mild quality regularizer, especially in highly biased regimes. By first analyzing the behavior of the fairness parameters under controlled conditions and then validating them on real submissions, Fair-PaperRec offers a practical, equity-focused framework for post-review paper selection that preserves—and in some settings can even enhance—measured scholarly quality.
\end{abstract}


\section{Introduction}
This journal version extends our earlier conference paper \cite{oyshi2025eknow}, offering a deeper and more comprehensive analysis of fairness-aware learning for mitigating post-review bias in academic peer review. Double-blind review often does not eradicate systemic biases linked to authors’ demographics, reputations, or institutional affiliations, despite attempts to ensure impartiality \cite{tomkins2017reviewer, shmidt2022double, giannakakos2025impact, mebane2025double}. Recent data indicate that even the most stringent anonymization techniques can be undermined by analyzing writing style or cross-referencing previous articles \cite{bauersfeld2023cracking, shah2023role}. This tendency can sustain biases against particular groups, including women, racial minorities, and researchers from underrepresented areas \cite{huber2022nobel, frachtenberg2022metrics, lee2013bias, giannakakos2025impact}. Simultaneously, there is a growing dependence on recommendation algorithms to optimize processes such as paper selection, grant distribution, and significant publication identification \cite{goues2017effectiveness, bobadillaDeepFair, peng2023reranking}. While these systems can accelerate decision-making, they also pose a danger of perpetuating biases present in the training data, particularly if they focus only on predictive accuracy \cite{morik2020controlling, beutel2017data, yao2017beyond}. Therefore, it is imperative to devise novel methodologies that explicitly include demographic justice, preventing the perpetuation of historical inequalities.

We start from a simple, testable hypothesis on synthetic data: increasing the strength of fairness regularization ($\lambda$) should improve inclusion with limited impact on utility, and there should exist a sweet-spot range that depends on the underlying bias level. We map this hypothesis across controlled regimes (fair, moderate, high bias) for multiple protected attributes (race, country), then carry the learned settings into \emph{original} conference data (ACM SIGCHI, DIS, IUI) to validate external relevance. This two-stage progression, from controlled what-if to real-world does-it-hold, reveals when fairness regularization uncovers under-selected high-quality work and when larger $\lambda$ values begin to over-correct a system that is already close to balanced.

In this paper, we introduce \emph{Fair-PaperRec}, a fairness-aware recommendation framework specifically designed to mitigate \emph{post-review} bias through a differentiable fairness loss integrated with a prediction objective. Unlike heuristic approaches that often handle single-attribute constraints or overlook intersectionality, our approach:

\begin{figure*}[ht]
    \captionsetup{font={footnotesize,rm},justification=centering,labelsep=period}%
    \centering
    \includegraphics[width=0.95\linewidth]{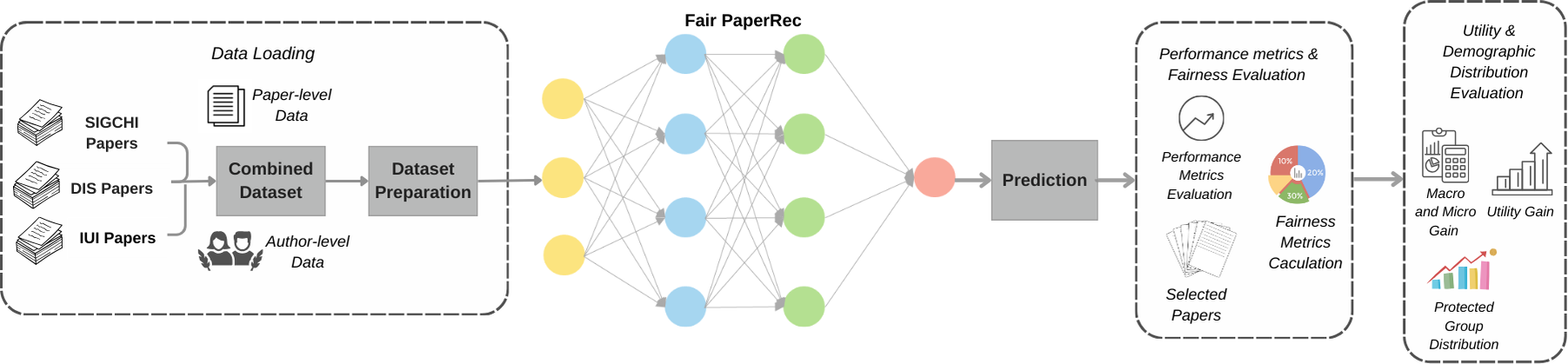}
    \caption{Overview of the Fair-PaperRec Architecture.}
    \label{fig:arch}
    \vskip -10pt
\end{figure*}
\captionsetup{font={footnotesize,rm},justification=centering,labelsep=period}%

\begin{itemize}
    \item Surpasses single-attribute approaches by incorporating multiple demographic attributes (e.g., race, country) and constructing profiles that capture underlying biases.
    \item After double-blind review, a specialized fairness penalty is implemented to address demographic disparities, correcting latent biases without replacing existing processes.
    \item The method explicitly targets demographic parity while monitoring utility, achieving equitable representation without compromising academic rigor.
\end{itemize}

Our results demonstrate improved representation in the participation of underrepresented groups, as well as an enhancement in overall paper quality. Notably, these findings reveal that enhanced inclusivity need not diminish academic rigor; a fairness-driven approach can yield greater demographic parity while simultaneously preserving, and at times even \emph{enhancing}, the quality of accepted papers.

\paragraph*{Contributions}
This paper makes three contributions:

\begin{enumerate}[leftmargin=*, itemsep=2pt, topsep=2pt]
    \item \textit{A post-review, multi-attribute fairness framework} that integrates a differentiable fairness loss with prediction to address intersectional disparities while preserving quality.
    \item \textit{A two-stage evaluation methodology} that (i) charts fairness-utility trade-offs on synthetic bias regimes to identify $\lambda$ sweet spots, then (ii) validates transfer to original conference data (SIGCHI, DIS, IUI).
    \item \textit{Actionable guidance} for choosing $\lambda$ and attributing weights based on disparity levels, showing when fairness delivers win-win improvements and when smaller regularization suffices.
\end{enumerate}

By mitigating biases in paper selection, our strategy promotes a richer academic discourse and amplifies the representation of marginalized communities, thereby paving the way toward more equitable, high-quality conferences. The paper is organized as follows: Section~II reviews related work; Section~III presents the methodology; Section~IV details the experimental setup and metrics; Section~V reports synthetic and original results with analysis; and Section~VI concludes.

\section{Related Work}
We review three strands of literature to motivate our two-stage story, starting from a simple hypothesis on synthetic data and ending in a real-world conclusion on original conference data. First, we examine where and why double-blind review can fail, emphasizing structural incentives beyond anonymity. Second, we connect these pressures to fairness in recommendation and ranking. Third, we cover post-review mitigation and neural approaches most relevant to our setting.

\subsection{Double-Blind Review, Structural Incentives, and Bias in Academic Selection}
Although double-blind review conceals identities \cite{tomkins2017reviewer,shmidt2022double,giannakakos2025impact}, it often fails to eliminate disparities by gender, race, or geography \cite{lee2013bias,abimbola2023geobias}. Re-identification via stylometry or linkage to prior work can undermine anonymity \cite{bauersfeld2023cracking}, while prestige effects still favor well-known institutions \cite{lee2013bias}. Consequently, underrepresented groups remain disadvantaged \cite{williams_national_2015}, and acceptance gaps persist \cite{huber2022nobel,frachtenberg2022metrics}.

Beyond anonymity, \emph{structural incentives} can shape outcomes. Conflicts of interest (COIs) and funding ties have become more visible as industry participation in ML/NLP expands. Recent large-scale audits report the prevalence and statistical correlates of industry involvement across venues and years \cite{hagendorff2023industry,bosten2025coi}, with complementary bibliometric evidence that industry and academia contribute differently across problem types, data assets, and deployment pathways \cite{kotseruba2021industryvision,liang2024complementary,farber2024companyimpact}. Co-authorship and collaboration networks also influence visibility and impact \cite{ortega2014coauthorship}, which can indirectly translate into acceptance advantages. Field norms further matter: in computing, the centrality of \emph{conferences} (deadlines, rapid iteration) versus \emph{journals} can shape incentives and review practices \cite{cacm2009confvjour,cacm2025fossilization}. Together, these factors motivate \emph{post-review} fairness controls that are sensitive to venue norms and network effects, not just anonymization fidelity.

Because peer selection often conflates contribution and credit, authorship order practices vary across subfields and team sizes \cite{authorplacement2022,stackexchangeAuthOrder}. Likewise, $h$-index is field- and seniority-dependent, with wide normative ranges and known limitations as a sole quality proxy \cite{samwellHindex,paperpileHindex}. These observations motivate our decision to (i) treat author-level “utility” as a \emph{weighted} $h$-index by career stage and (ii) evaluate fairness at \emph{paper} and \emph{author} levels (macro/micro).

\subsection{Fairness in Recommendation and Ranking}
When optimizing solely for accuracy, recommenders can amplify historical bias \cite{burke2017multisided,bobadillaDeepFair}. Multi-objective \cite{morik2020controlling}, adversarial \cite{beutel2017data}, and re-ranking approaches \cite{yao2017beyond} offer alternatives but frequently target a \emph{single} protected attribute or generic user–item contexts, leaving intersectional and post-review constraints underexplored. In scholarly selection, \emph{provider fairness} aligns with \emph{author fairness}; the focus is on equitable exposure and acceptance for authors, not only on predictive accuracy \cite{alsaffar2021multidimensional}. Contemporary venue trends (e.g., topic shifts and data/benchmark access highlighted at CVPR 2024) underscore how exposure and resources interact with selection signals \cite{adobecvpr2024trends}, reinforcing the need to treat fairness and quality jointly.

In recommender-systems literature, \emph{provider fairness} refers to ensuring equitable exposure or selection opportunities for content creators. In the academic peer-review context, the “providers” are the authors themselves; thus, provider fairness corresponds directly to \emph{author fairness}, aiming to equalize acceptance probabilities across demographic groups. This distinction clarifies that our work focuses on fairness for authors (providers), rather than fairness for users of the recommendation system.

Our synthetic-to-original narrative builds on this literature by (i) explicitly modeling \emph{multiple} protected attributes (race, country), (ii) mapping fairness-utility trade-offs across tunable bias regimes to locate $\lambda$ “sweet spots,” and (iii) transferring those settings to real conference data where industry/academia dynamics, network effects, and venue norms co-exist \cite{kotseruba2021industryvision,liang2024complementary,farber2024companyimpact,cacm2009confvjour,cacm2025fossilization}.

\subsection{Post-Review Bias Mitigation and Neural Approaches}
Heuristic post-review balancing can improve representation \cite{alsaffar2021multidimensional} but risks local optima and often omits multi-attribute fairness. Neural approaches such as \emph{DeepFair} \cite{bobadillaDeepFair} and \emph{Neural Fair Collaborative Filtering} \cite{islam2021neural} show that fairness and accuracy can align in commercial domains, yet peer review imposes hard quotas, reputational stakes, and limited items. Multi-stakeholder optimization \cite{wu2022multi, wang2023survey} argues for context-aware definitions (e.g., exposure vs.\ acceptance parity), while text-based relevance features (e.g., TF–IDF) can help ranking \cite{bulut2018paper} but do not, on their own, guarantee equity for historically marginalized authors.

We focus on a \emph{post-review} neural re-ranking stage with a differentiable fairness loss that handles \emph{intersectional} attributes. The synthetic analysis provides a controlled test of the hypothesis that increasing fairness regularization improves inclusion with limited utility cost—and reveals regime-dependent sweet spots. The original-data validation then asks whether these settings hold under real constraints shaped by industry–academia participation, collaboration networks, and conference practices \cite{bosten2025coi,hagendorff2023industry,liang2024complementary,ortega2014coauthorship,cacm2009confvjour}. This two-stage path converts a simple, testable idea into a practical recipe for equity-preserving paper selection.

\section{Methodology}

Our approach tackles demographic biases in conference data to enforce fairness post-review: we begin with a simple, controlled hypothesis tested on synthetic datasets, then validate its robustness on original conference data. The design therefore emphasizes both (i) tractability in controlled regimes and (ii) fidelity to the complex realities of peer review. The backbone of our approach is a lightweight Multilayer Perceptron (MLP) trained with an integrated fairness loss, as illustrated in Fig.~\ref{fig:arch}. This choice reflects two guiding principles: (1) expose and alleviate demographic disparities instead of concealing them, and (2) use a simple, transparent neural architecture that harmonizes fairness and utility while remaining interpretable enough to trace effects across bias regimes.

\subsection{Data Collection and Pre-processing}

\paragraph{Synthetic stage (hypothesis testing)}
Synthetic datasets are constructed with tunable demographic distributions (fair, moderate, high bias). These provide a sandbox where the effect of fairness regularization $\lambda$ can be precisely measured. By varying the demographic skew, we can test the hypothesis that fairness constraints are most effective when initial bias is high, and that “sweet spot” $\lambda$ values differ by attribute. Table~\ref{table:synthetic_demographics} summarizes the demographic distributions for each protected group across the fair, moderate, and high synthetic bias scenarios.

\paragraph{Original stage (validation)}
To evaluate transfer to real contexts, we utilize datasets from SIGCHI 2017, DIS 2017, and IUI 2017~\cite{alsaffar2021multidimensional}. These venues naturally exhibit systemic imbalances in demographics and prestige, reflecting the challenges of real peer review. Rather than discarding these biases, we preserve them to observe how fairness regularization reshapes outcomes in situ. Table~\ref{table:demographics} reports the participation rates of protected groups (gender, race, country) in the three conferences (SIGCHI, DIS, IUI).

\paragraph{Processing pipeline}
At both stages, the dataset is curated at paper and author levels. Every paper record includes a title, author list, and conference label (1 = IUI, 2 = DIS, 3 = SIGCHI). Author records capture demographic variables (gender, race, nationality, career stage). 
Real-world datasets—particularly those drawn from academic conference submissions—often contain latent biases that mirror systemic imbalances in the scholarly community (e.g., underrepresentation of certain demographics). We utilize datasets from conferences that naturally reflect systemic disparities (e.g., skewed demographics). Instead of eliminating such biases, our objective is to \emph{recognize and rectify} them.

\begin{table}
\centering
\caption{Demographic Distribution in Synthetic Datasets Across Bias Scenarios For Protected Groups}
\label{table:synthetic_demographics}
\begin{tabular}{lccc}
\toprule
\textbf{Scenario} & \textbf{Gender (\%)} & \textbf{Race (\%)} & \textbf{Country (\%)} \\
\midrule
Fair     & 48.80       & 51.50     & 52.30        \\
Moderate & 28.10       & 28.90     & 31.50        \\
High     & 8.50        & 9.70      & 10.00        \\
\midrule
Average  & 28.47       & 30.03     & 31.27        \\
\bottomrule
\end{tabular}
\end{table}

\begin{table}[ht]
\centering
\captionsetup{font={footnotesize,rm},justification=centering,labelsep=period}%
\caption{\MakeUppercase{Demographic Participation from protected groups in Three Conferences.}}
\label{table:demographics}
\begin{tabular}{lccc}
    \toprule
    \textbf{Conference} & \textbf{Gender (\%)} & \textbf{Race (\%)} & \textbf{Country (\%)} \\
    \midrule
    SIGCHI & 41.88 &   6.84 &    21.94 \\
    DIS    & 65.79 &  35.09 &    24.56 \\
    IUI    & 43.75 &  51.56 &    39.06 \\
    \midrule
    Average & 50.47 &  31.16 &    28.52 \\
    \bottomrule
\end{tabular}
\end{table}
\captionsetup{font={footnotesize,rm},justification=centering,labelsep=period}%

\subsubsection{Data Description}
The resulting dataset combines both paper-level and author-level perspectives. \emph{Overall} refers to all submissions, while \emph{Selected} refers to those recommended by Fair-PaperRec. SIGCHI 2017 papers act as a benchmark for high-impact work, offering a ground truth for quality. Author records include career stages (student, postdoc, faculty, industry), enabling $h$-index weighting that reflects differences in scholarly maturity.

\subsubsection{Data Pre-processing}
Several preprocessing steps were undertaken to prepare the dataset for training:

    \emph{Categorical Encoding:} Gender, Country, and Race are subjected to one-hot encoding. Gender is binary (0 = male, 1 = female), Country is categorized as \emph{developed} or \emph{underdeveloped}, and Race comprises \{White, Asian, Hispanic, Black\}, with Hispanic and Black designated as protected groups (Table~\ref{table:demographics}).
    \emph{Normalization:} Numerical attributes (e.g., h-index) employ min-max scaling for consistent magnitude.
    \emph{Training and Validation Division:} An 80\%/20\% stratified division guarantees equitable distribution of labels and protected attributes in both subsets.

\subsection{Problem Definition}

This study develops a \emph{fairness-aware paper recommendation system} that ensures demographic parity with respect to authors’ race and country while preserving high academic standards. We frame acceptance decisions as a \emph{recommendation} task, where \emph{conference organizers (users)} seek to select from \emph{530 papers (items)} spanning SIGCHI, DIS, and IUI. Each paper (item) includes an \emph{h-index} for quality, demographic data (race, country), and a conference rating.

Our approach enforces fairness constraints on race and country independently, excluding \emph{gender} due to its relatively balanced distribution in the dataset (Table~\ref{table:demographics}). Preliminary analysis further showed minimal disparity between accepted and rejected papers with respect to gender; because our fairness framework penalizes deviations from the observed baseline, incorporating gender would have produced negligible changes in the fairness–utility trade-offs examined here. To maintain clarity of exposition and focus on attributes with substantially higher disparity, we therefore restricted fairness constraints to race and country. Nevertheless, gender remains an important dimension of equity, and future work will extend Fair-PaperRec to gender-aware and multi-attribute fairness in larger or more imbalanced datasets.

Let \( D \) represent the dataset of submitted papers, where each paper \( p \in D \) is associated with a set of features \( X_p \) (e.g., race, country, h-index) and a target variable \( y_p \) indicating acceptance (1) or rejection (0). The \emph{race} attribute \( R_p \) and \emph{country} attribute \( C_p \) are the protected attributes.

We aim to optimize a predictive model \( f: X_p \rightarrow \hat{y}_p \) that minimizes the following objective function:

\begin{equation}
    \min_{f} \left( \mathcal{L}(f(X_p), y_p) + \lambda \cdot \mathcal{L}_{\text{fairness}}(f, D) \right)
    \label{eq:eq_1}
\end{equation}

Here, \( \mathcal{L}(f(X_p), y_p) \) is the \emph{prediction loss} (e.g., Binary Cross-Entropy Loss), \( \mathcal{L}_{\text{fairness}}(f, D) \) is the \emph{fairness loss}, penalizing deviations from demographic parity across race and country, and \( \lambda \) is a hyperparameter that balances the trade-off between prediction accuracy and fairness.

\begin{algorithm}
\captionsetup{font={footnotesize,rm},justification=centering,labelsep=period}%
\caption{\MakeUppercase{Fair-PaperRec Loss Function.}}
\footnotesize
\label{alg:algorithm1}
\begin{algorithmic}[1]
    \State \textbf{Input}: Model \( M \), Epochs \( E \), Batch size \( B \), Data \( D \), Protected attributes \( A \), Hyperparameter \( \lambda \)
    \State \textbf{Output}: Trained Model \( M \)
    \State \textbf{Initialize} Model \( M \)
    \For{each \( e \in E \)}
        \State Shuffle Data \( D \)
        \For{each batch \( \{ (X, Y) \} \in D \) with size \( B \)}
            \State Predict \( \hat{Y} \leftarrow M(X) \)
            \State Calculate Loss:
            \State \quad \( L_{\text{prediction}} \leftarrow \text{PredictionLoss}(Y, \hat{Y}) \)
            \State \quad \( L_{\text{fairness}} \leftarrow \text{FairnessLoss}(A, \hat{Y}) \)
            \State Calculate Total Loss:
            \State \quad \( L_{\text{total}} \leftarrow \lambda \cdot L_{\text{fairness}} + L_{\text{prediction}} \)
            \State Compute gradients \( \nabla L_{\text{total}} \leftarrow \frac{\partial L_{\text{total}}}{\partial M} \)
            \State Update Model parameters: \( M \leftarrow M - \alpha \nabla L_{\text{total}} \)
        \EndFor
    \EndFor
\end{algorithmic}
\end{algorithm}
\captionsetup{font={footnotesize,rm},justification=centering,labelsep=period}%

\subsection{Demographic Parity}
We aim to ensure that the probability of a paper being accepted is independent of the protected attributes:
\begin{equation*}
    P(\hat{y}_p = 1 \mid R_p = r) = P(\hat{y}_p = 1), \quad \forall r \in \text{Race}
    \label{eq:eq_2}
\end{equation*}
\begin{equation*}
    P(\hat{y}_p = 1 \mid C_p = c) = P(\hat{y}_p = 1), \quad \forall c \in \text{Country}
    \label{eq:eq_3}
\end{equation*}

Utilizing these equations ensures that the papers authored by individuals from different races and countries have an equal probability of acceptance.

\subsection{Fairness Loss}
The fairness loss from the objective function in Equation ~(\ref{eq:eq_1}) is constructed to minimize statistical parity differences between the protected and non-protected groups:
\begin{equation}
    \mathcal{L}_{\text{fairness}} = \left( P(\hat{y}_p = 1 \mid G_{\text{p}}) - P(\hat{y}_p = 1 \mid G_{\text{np}}) \right)^2
    \label{eq:eq_4}
\end{equation}

Here, \( P(\hat{y}_p = 1 \mid G_{\text{p}}) \) denotes the acceptance probability for the protected group and \( P(\hat{y}_p = 1 \mid G_{\text{np}}) \) is the acceptance probability for the non-protected group.

\subsection{Combined Fairness Loss}
Furthermore, we define a combined fairness loss to minimize statistical parity differences across race and country attributes between the protected and unprotected groups, as shown in Equation ~(\ref{eq:combinedloss}).

\begin{equation}
\begin{split}
    \mathcal{L}_{\text{fairness}} = W_{r} 
    \left( \frac{1}{N_{r}} \sum_{p \in G_{r}} \hat{y}_p 
    - \frac{1}{N} \sum_{p=1}^{N} \hat{y}_p \right)^2 \\
    + W_{c} 
    \left( \frac{1}{N_{c}} \sum_{p \in G_{c}} \hat{y}_p 
    - \frac{1}{N} \sum_{p=1}^{N} \hat{y}_p \right)^2
\end{split}
\label{eq:combinedloss}
\end{equation}
\( G_{\text{r}} \) and \( G_{\text{c}} \) denote the race and country groups, respectively.
\( N_{r} \) and \( N_{c} \) are the number of papers in each group and weights \( W_{r} \) and \( W_{c} \) reflect group distributions.

\subsection{Total Loss}
The total loss is the combination of prediction and fairness losses:

\begin{equation*}
    \mathcal{L}_{\text{total}} = \mathcal{L}_{\text{prediction}} + \lambda \cdot \mathcal{L}_{\text{fairness}}
\end{equation*}

\subsection{Constraints and Considerations}
We assess fairness by training our model separately on \emph{race} and \emph{country}, as well as jointly on both attributes to evaluate selection fairness across multiple dimensions.

\paragraph{Exclusion of Protected Attributes}
Race \(R_p\) and country \(C_p\) are excluded from the input feature set \(X_p\) to mitigate direct bias amplification. To achieve joint fairness, both attributes are omitted during training, preventing the model from learning acceptance outcomes influenced by race or country.

\paragraph{Indirect Bias Mitigation}
A fairness loss promotes demographic parity, addressing indirect biases associated with features related to race or country. The model maintains neutrality by penalizing selection disparities, even in the absence of protected attributes.

\paragraph{Scalability}
The MLP architecture is deliberately simple, enabling reproducibility across venues and bias levels. Synthetic results provide controlled insight into when fairness is most beneficial; original results demonstrate external validity. Together, they build a methodological bridge from a simple hypothesis (“fairness regularization should help most under bias”) to a powerful conclusion (“in real conferences, tuning $\lambda$ uncovers under-selected, high-quality work without compromising rigor”).

\section{Model Overview}

To achieve demographic parity while preserving quality in paper selection, we present a MLP-based neural network (see Fig. \ref{fig:arch}), explicitly engineered to balance the trade-off between fairness and accuracy. It illustrates the correlations between input features, like author demographic attributes and paper quality, while alleviating biases during selection.

A unique fairness loss function was employed to ensure equity, imposing penalties on the model for substantial differences in selection rates between protected and non-protected groups. This loss function is integrated with the conventional prediction loss to attain a balance between diversity and accuracy; the algorithm is shown in Algorithm \ref{alg:algorithm1}.

The acceptance probabilities for submitted papers are generated by the MLP, which are subsequently ranked to guarantee that the final selection meets both quality and fairness objectives. By selecting top papers according to these probabilities, we ensure equal representation of authors from both protected and non-protected groups while upholding the requisite standard of academic excellence.

\begin{algorithm}
\captionsetup{font={footnotesize,rm},justification=centering,labelsep=period}%
\caption{\MakeUppercase{Fairness-Aware Paper Selection Mechanism.}}
\footnotesize
\label{alg:algo2}
\begin{algorithmic}[1]
    \State \textbf{Input}: Dataset \( D \), Model \( M \), Number of Accepted Papers \( N_a \), Total Papers \( N_t \)
    \State \textbf{Output}: Selected Papers \( P_{\text{selected}} \)
    \State \textbf{Initialize}: \( P_{\text{selected}} \leftarrow \emptyset \)
    \State \textbf{Step 1}: Apply trained model \( M \) to the entire dataset \( D \)
    \For{each paper \( p \in D \)}
        \State Compute acceptance probability: \( \hat{y}_p \leftarrow M(p) \)
    \EndFor
    \State \textbf{Step 2}: Rank all papers \( p \) by acceptance probability \( \hat{y}_p \)
    \State Sort \( D \) in descending order of \( \hat{y}_p \)
    \State \textbf{Step 3}: Select top \( N_a \) papers:
    \State \quad \( P_{\text{selected}} \leftarrow \{ p \mid \hat{y}_p \geq \hat{y}_{(N_a)} \} \)
    \State \textbf{Step 4}: Ensure Fairness Constraints
    \State \textbf{Return} \( P_{\text{selected}} \)
\end{algorithmic}
\end{algorithm}
\captionsetup{font={footnotesize,rm},justification=centering,labelsep=period}%
\vskip -10pt

\begin{table*}[ht]
\centering
\captionsetup{font={footnotesize,rm},justification=centering,labelsep=period}%
\caption{\MakeUppercase{Relative Gain (\%) for Macro, Micro, and Utility Metrics across $\lambda \in [1,10]$ for Country and Race Features under Different Bias Levels.}}
\label{table:relative_gain_side}
\renewcommand{\arraystretch}{1.05}
\setlength{\tabcolsep}{4pt}
\scriptsize
\begin{tabular}{ll ccc ccc}
    \toprule
    & & \multicolumn{3}{c}{\textbf{Country Feature}} & \multicolumn{3}{c}{\textbf{Race Feature}} \\
    \cmidrule(lr){3-5} \cmidrule(lr){6-8}
    \textbf{Bias Level} & $\boldsymbol{\lambda}$ & \makecell{\textbf{Macro}\\\textbf{Gain (\%)}} & \makecell{\textbf{Micro}\\\textbf{Gain (\%)}} & \makecell{\textbf{Utility}\\\textbf{Gain (\%)}} & \makecell{\textbf{Macro}\\\textbf{Gain (\%)}} & \makecell{\textbf{Micro}\\\textbf{Gain (\%)}} & \makecell{\textbf{Utility}\\\textbf{Gain (\%)}} \\
    \midrule

    \multirow{5}{*}{High}
    & 1   & -18.03 & -4.17 & -2.98 & \textbf{15.03} & 20.18 & 8.51 \\
    & 2.5 & -3.93  & -1.65 & -1.28 & 12.44 & 18.46 & 2.55 \\
    & 3   & -1.18  &  3.22 &  5.96 & 12.11 & \textbf{21.10} & \textbf{15.32} \\
    & 5   & -0.30  & -2.61 &  8.09 & 10.12 & 16.84 & 14.47 \\
    & 10  & \textbf{4.98}  & \textbf{9.22} & \textbf{10.21} & 5.06  & 14.20 & 5.53 \\
    \midrule

    \multirow{5}{*}{Moderate}
    & 1   &  2.90 & \textbf{10.44} & \textbf{15.32} & 3.00 & \textbf{10.62} & 11.06 \\
    & 2.5 & -1.11 & -2.83 & 4.68 & 1.82 & 4.66  & 8.09 \\
    & 3   & \textbf{5.04} & 3.97  & 7.23 & \textbf{3.96} & 9.54  & \textbf{13.62} \\
    & 5   & -3.15 & -6.15 & 0.43 & 4.39 & 10.55 & 11.06 \\
    & 10  &  2.29 &  0.49 & -0.85 & 0.83 & 3.05  & 7.66 \\
    \midrule

    \multirow{5}{*}{Fair}
    & 1   & \textbf{5.35} & \textbf{7.26} & \textbf{20.43} & -6.51 & 0.79 & -4.26 \\
    & 2.5 & 0.51  & 0.02 & 8.94 & -0.28 & -0.30 & 11.06 \\
    & 3   & 3.81  & 2.01 & -0.85 & -8.96 & -12.78 & 8.09 \\
    & 5   & 2.31  & 1.43 & 3.83 & -2.12 & -4.76 & \textbf{12.77} \\
    & 10  & 3.93  & 2.12 & 10.64 & \textbf{1.37} & -1.78 & 4.68 \\
    \bottomrule
\end{tabular}
\vskip -8pt
\end{table*}

\subsection{Selection Mechanism}

The model calculates acceptance probabilities for all submitted papers after training. After calculating acceptance odds, the algorithm ranks candidate papers. This rating phase ensures that underrepresented groups are represented in the final admission decisions. Representing this as a suggestion list preserves the peer-review process and corrects residual biases. Algorithm \ref{alg:algo2} selects the best papers based on probability, ensuring fairness and preserving the desired number of accepted papers.
\begin{itemize}
    \item \emph{Prediction Aggregation:} The trained MLP model is applied to the entire dataset to obtain predicted acceptance probabilities \( \hat{y}_p \) for each paper. 
    \item \emph{Ranking:} Papers are ranked in descending order based on their predicted probabilities.
    \item \emph{Selection:} The papers with the highest predicted probabilities are selected for acceptance, ensuring that the total number of selected papers matches the required acceptance quota.
\end{itemize}

Mathematically, the selection process is represented as:
\begin{equation*}
\text{Selected Papers} = \left\{ p \in D \mid \hat{y}_p \geq \hat{y}_{(N_a)} \right\}
\end{equation*}

Here, \( \hat{y}_{(N_a)} \) is the \( N_a \)-th highest predicted probability in the set \( \{ \hat{y}_p \mid p \in D \} \) while \( N_a \) is the total number of accepted papers and \( N_t \) is the total number of submitted papers, where \( N_a \leq N_t \).

This approach ensures that the selection process is both informed by the model's predictions and constrained to uphold demographic parity, fostering an equitable and meritocratic paper selection environment.

\section{Evaluation and Experiments}
This section presents the experimental evaluation of our proposed Fair-PaperRec model on the chosen datasets. First, we test our hypothesis in controlled synthetic settings, then we validate it with original conference data. This design allows us to ask not only whether fairness interventions work, but also \emph{when} and \emph{why} they are most effective.

\begin{figure*}[t]
\centering

\subfloat[Race — Fair bias]{%
  \includegraphics[width=0.32\linewidth]{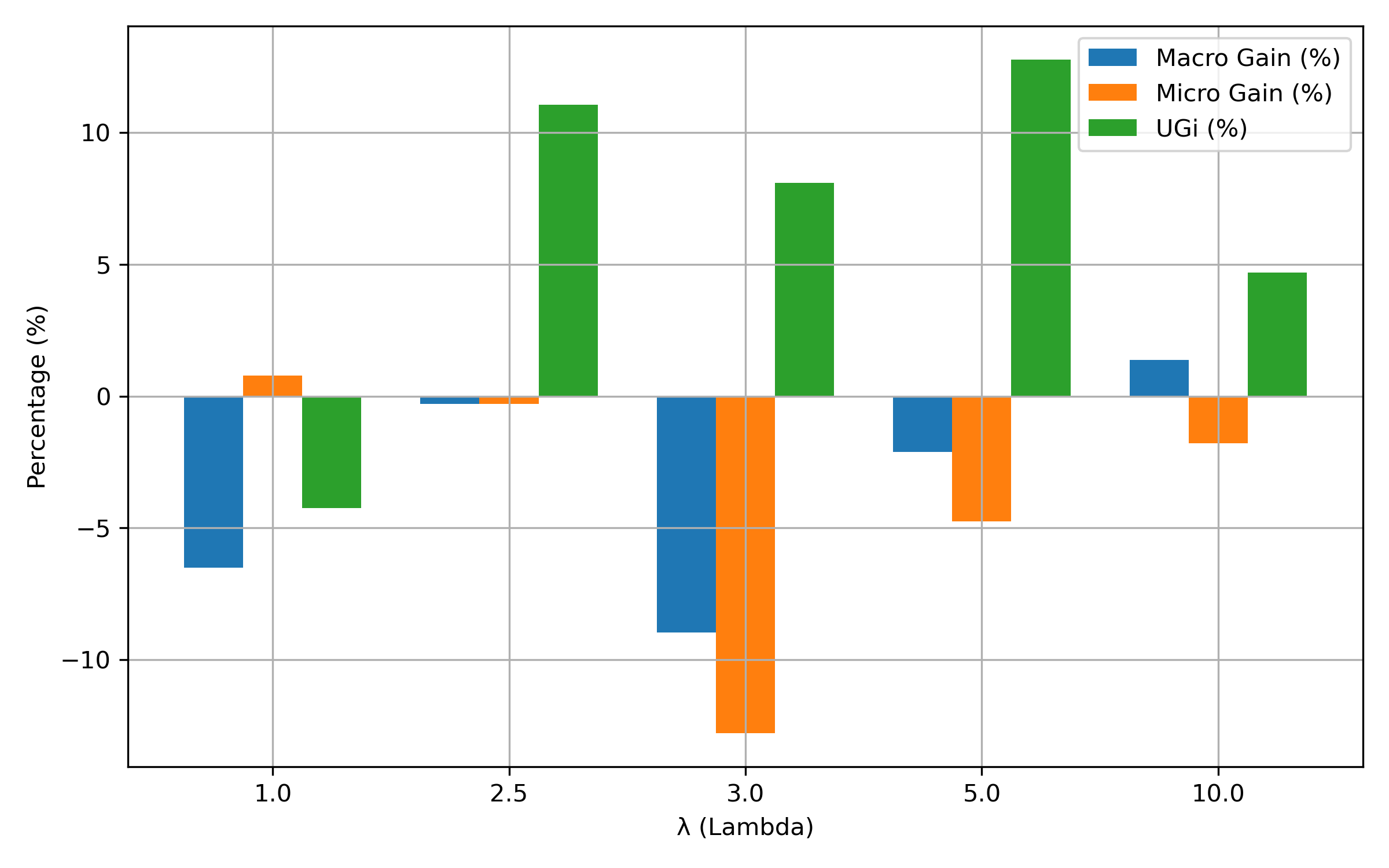}
}\hfill
\subfloat[Race — Moderate bias]{%
  \includegraphics[width=0.32\linewidth]{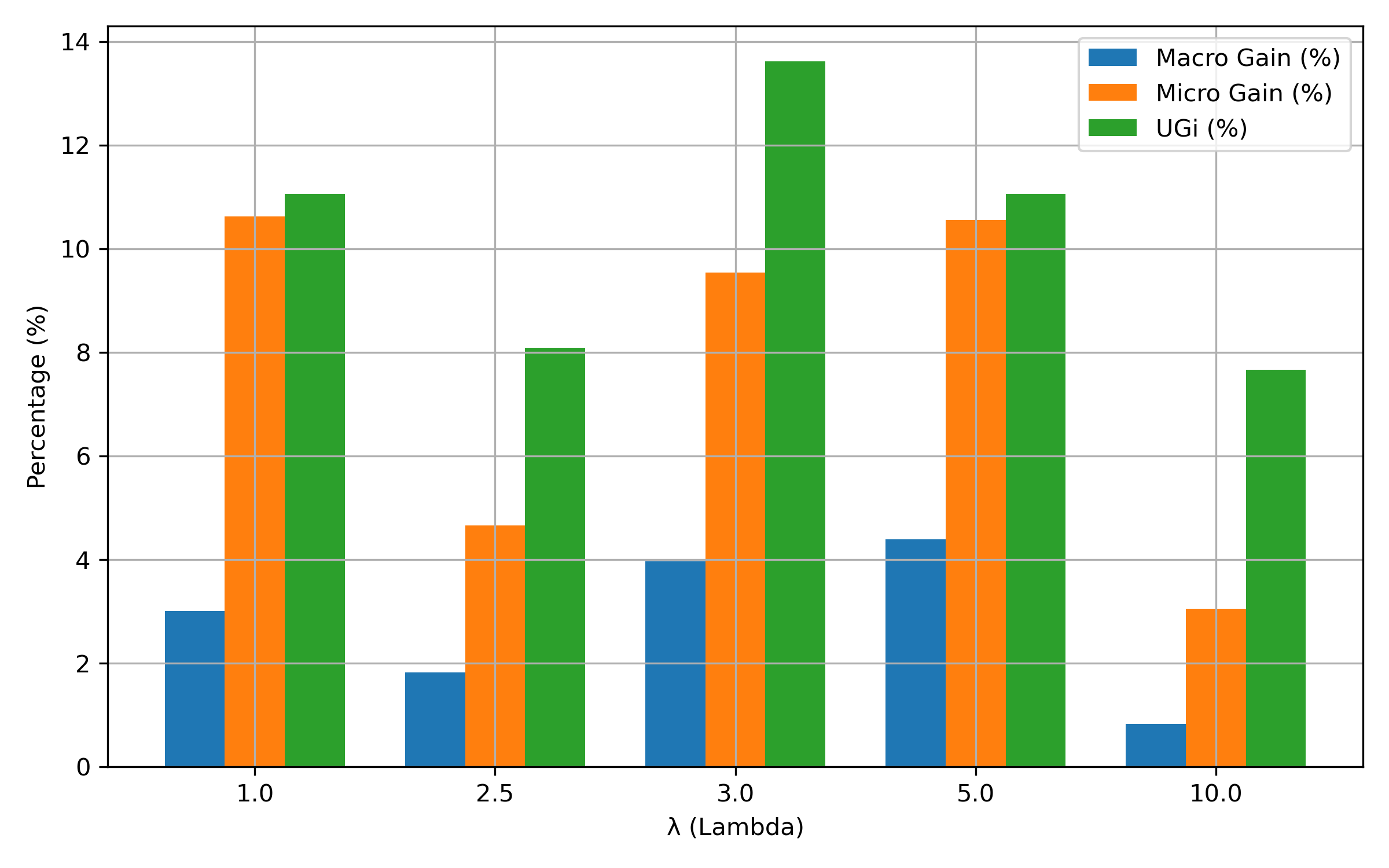}
}\hfill
\subfloat[Race — High bias]{%
  \includegraphics[width=0.32\linewidth]{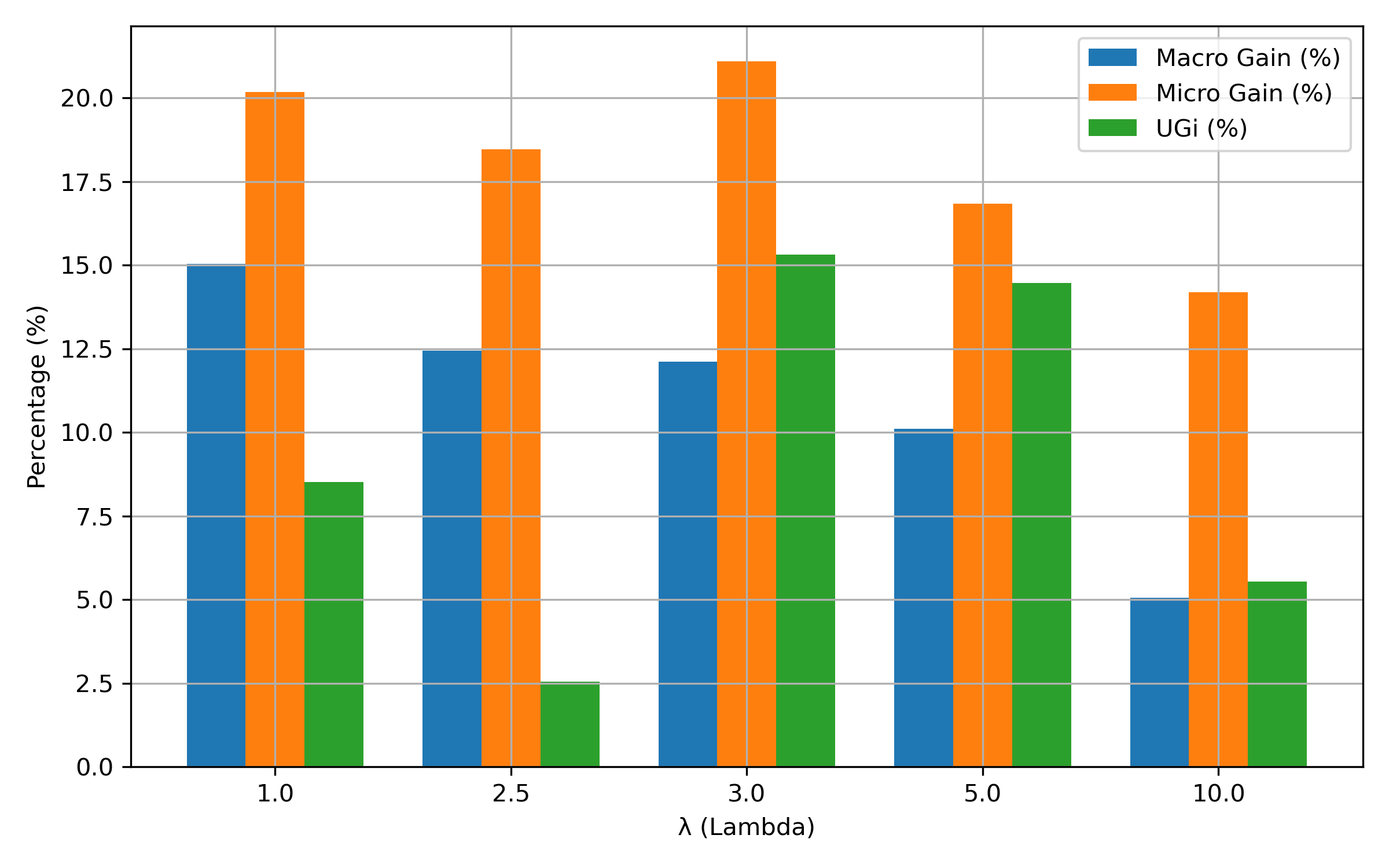}
}

\vspace{0.6em}

\subfloat[Country — Fair bias]{%
  \includegraphics[width=0.32\linewidth]{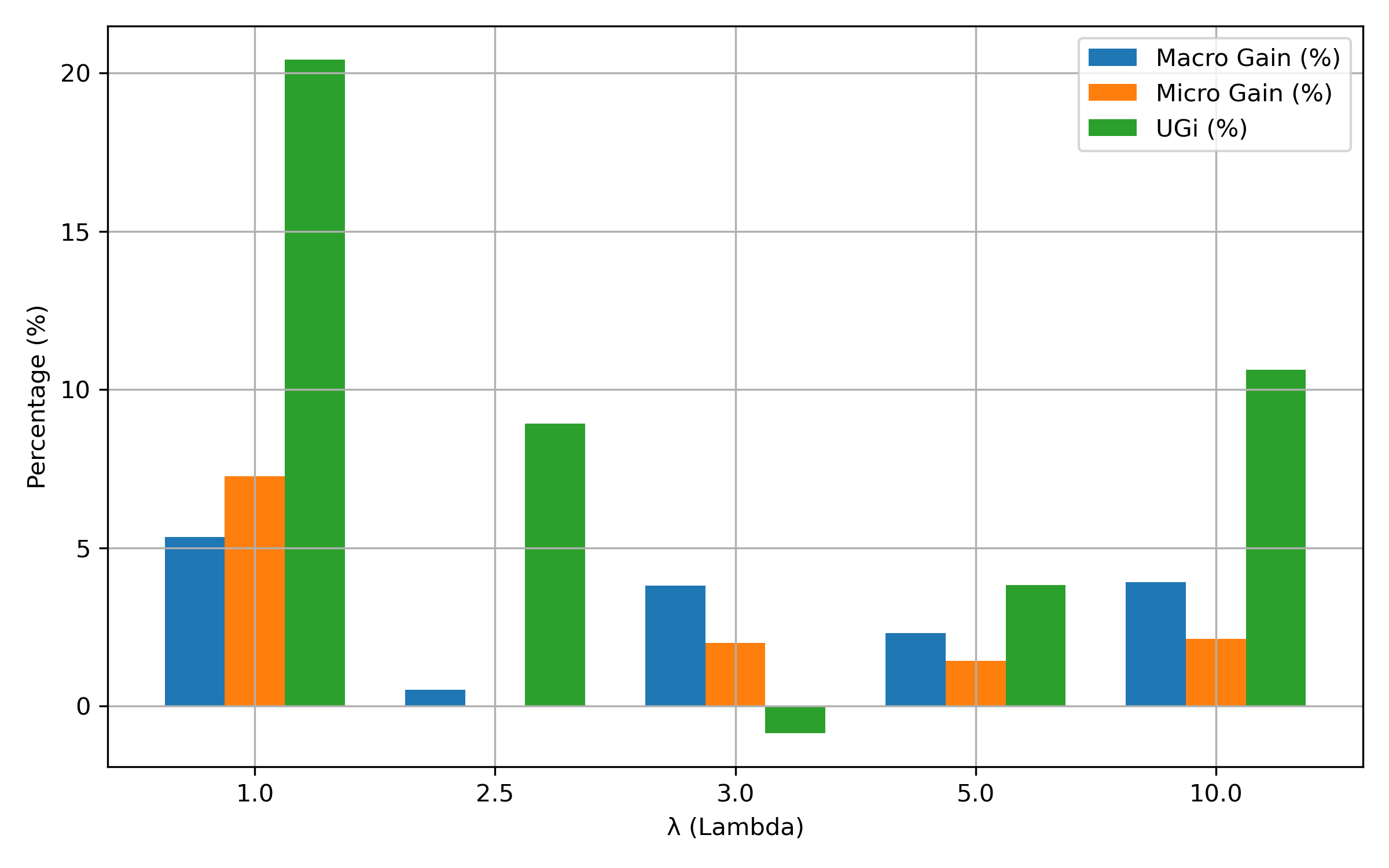}
}\hfill
\subfloat[Country — Moderate bias]{%
  \includegraphics[width=0.32\linewidth]{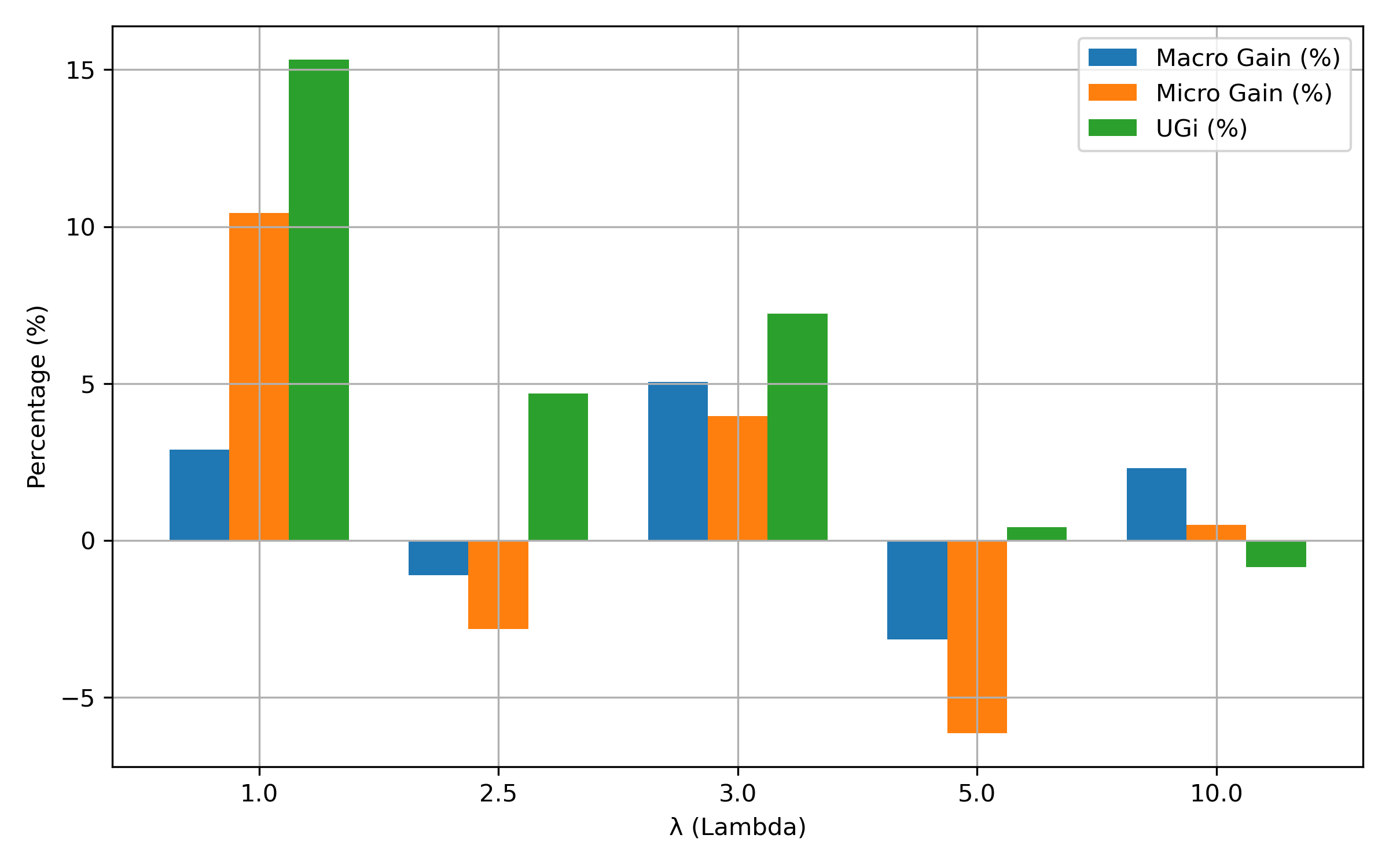}
}\hfill
\subfloat[Country — High bias]{%
  \includegraphics[width=0.32\linewidth]{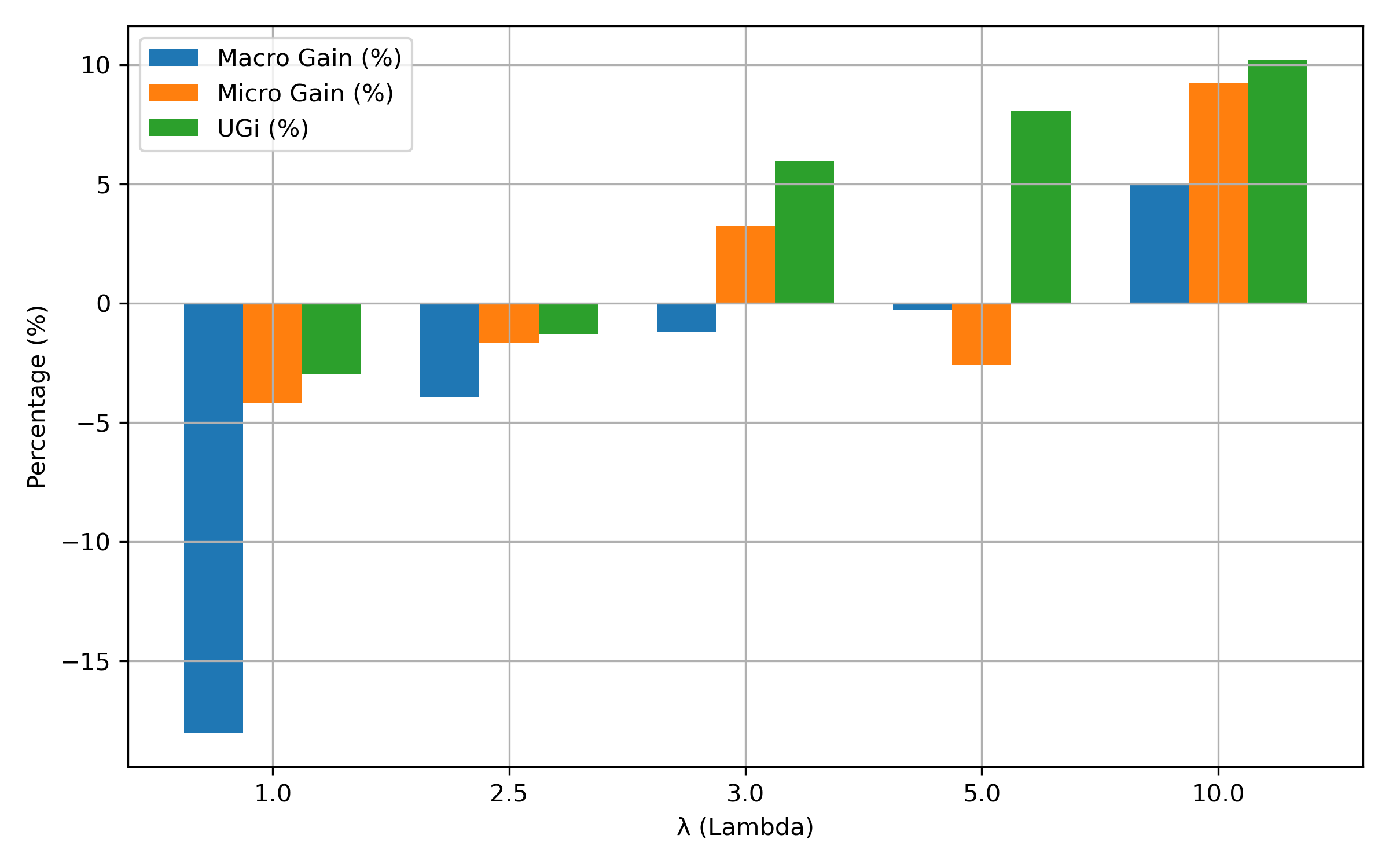}
}

\caption{Relative gains (\%) for Macro, Micro, and Utility across fairness regularization strengths ($\lambda\in[1,10]$). Top row: Race; bottom row: Country.}
\label{fig:gain_grid}
\end{figure*}

To guide this exploration, we pose three research questions:

\begin{itemize}[leftmargin=*]
    \item \emph{RQ1:} How do fairness constraints affect the overall quality (utility) of recommended papers, as measured by author $h$-index?
    \item \emph{RQ2:} Does handling race and country as separate protected attributes differ from treating them jointly in terms of fairness outcomes and selection decisions?
    \item \emph{RQ3:} How do varying weight assignments to multiple protected attributes (race and country) influence the trade-off between fairness and utility?
\end{itemize}

These questions move us along the synthetic-to-original trajectory, from controlled hypothesis testing to real-world validation. RQ1 begins in the synthetic environment, where we test the basic hypothesis that fairness regularization ($\lambda$) improves inclusion with limited cost to utility. RQ2 and RQ3 extend to the real-world stage, asking whether the same patterns hold under authentic demographic distributions and multi-attribute interactions.

\subsection{Experimental Setting}
We evaluate Fair-PaperRec across both synthetic datasets (fair, moderate, and high-bias regimes) and original conference datasets. This dual evaluation ensures that our findings generalize from controlled testbeds to complex real-world settings where structural factors (prestige, career stage, institutional affiliation) interact with demographics.

Each experiment is conducted five times with independent random seeds. We report mean values with \emph{standard deviations} to highlight stability across runs, ensuring that observed peaks or trade-offs are not artifacts of stochastic training.

\subsubsection{Implementation Details}
All experiments are implemented in \texttt{PyTorch} and executed on a high-performance server equipped with two NVIDIA Quadro RTX 4000 GPUs. The Fair-PaperRec model is a two-hidden-layer MLP using Rectified Linear Unit (ReLU) activations and Batch Normalization, culminating in a sigmoid output layer that produces acceptance probabilities. 
\begin{table}[h!]
\centering
\captionsetup{font={footnotesize,rm},justification=centering,labelsep=period}%
\caption{\MakeUppercase{Distribution of Recommended Papers from Each Tier for Synthetic Dataset.}}
\label{table:synth}
\begin{tabular}{lccc}
\toprule
\textbf{Label} & \textbf{Country} & \textbf{Race} & \textbf{Multi-Fair} \\
\midrule
Top-tier  & 87.12\% & 89.30\% & 91.41\% \\
Mid-tier     & 6.45\%  & 6.87\%  & 5.64\% \\
Low-tier    & 6.43\%  & 3.83\%  & 2.95\% \\
\midrule
\textbf{\# Papers} & 280 & 280 & 280 \\
\bottomrule
\end{tabular}
\vskip -10pt
\end{table}
\captionsetup{font={footnotesize,rm},justification=centering,labelsep=period}%

We train for 50 epochs using the Adam optimizer (learning rate = 0.001). Early stopping is applied if validation performance does not improve over 10 consecutive epochs. The fairness regularization parameter $\lambda$ is swept across a range of values (1–10) to trace fairness–utility trade-offs. Datasets are split 80/20 (training/validation) using stratified sampling, preserving both label and demographic distributions.

This setup allows us to simulate the “synthetic hypothesis-testing stage” (controlled bias ratios) and then apply the same procedure in the “original validation stage” (real demographic skew).

\begin{table}[h!]
\centering
\captionsetup{font={footnotesize,rm},justification=centering,labelsep=period}%
\caption{\MakeUppercase{Distribution of Recommended Papers from Each Conference.}}
\label{table:conf_dist}

\begin{tabular}{lccc}
\toprule
\textbf{Label} & \textbf{Country} & \textbf{Race} & \textbf{Multi-Fair} \\
\midrule
SIGCHI  & 92.02\% & 92.00\% & 92.02\% \\
DIS     & 4.84\%  & 7.69\%  & 7.40\% \\
IUI     & 3.14\%  & 0.31\%  & 0.56\% \\
\midrule
\textbf{\# Papers} & 351 & 351 & 351 \\
\bottomrule
\end{tabular}
\vskip -10pt
\end{table}
\captionsetup{font={footnotesize,rm},justification=centering,labelsep=period}%

\begin{figure}[ht]
    \captionsetup{font={footnotesize,rm},justification=centering,labelsep=period}%
    \centering
    \includegraphics[width=0.9\linewidth]{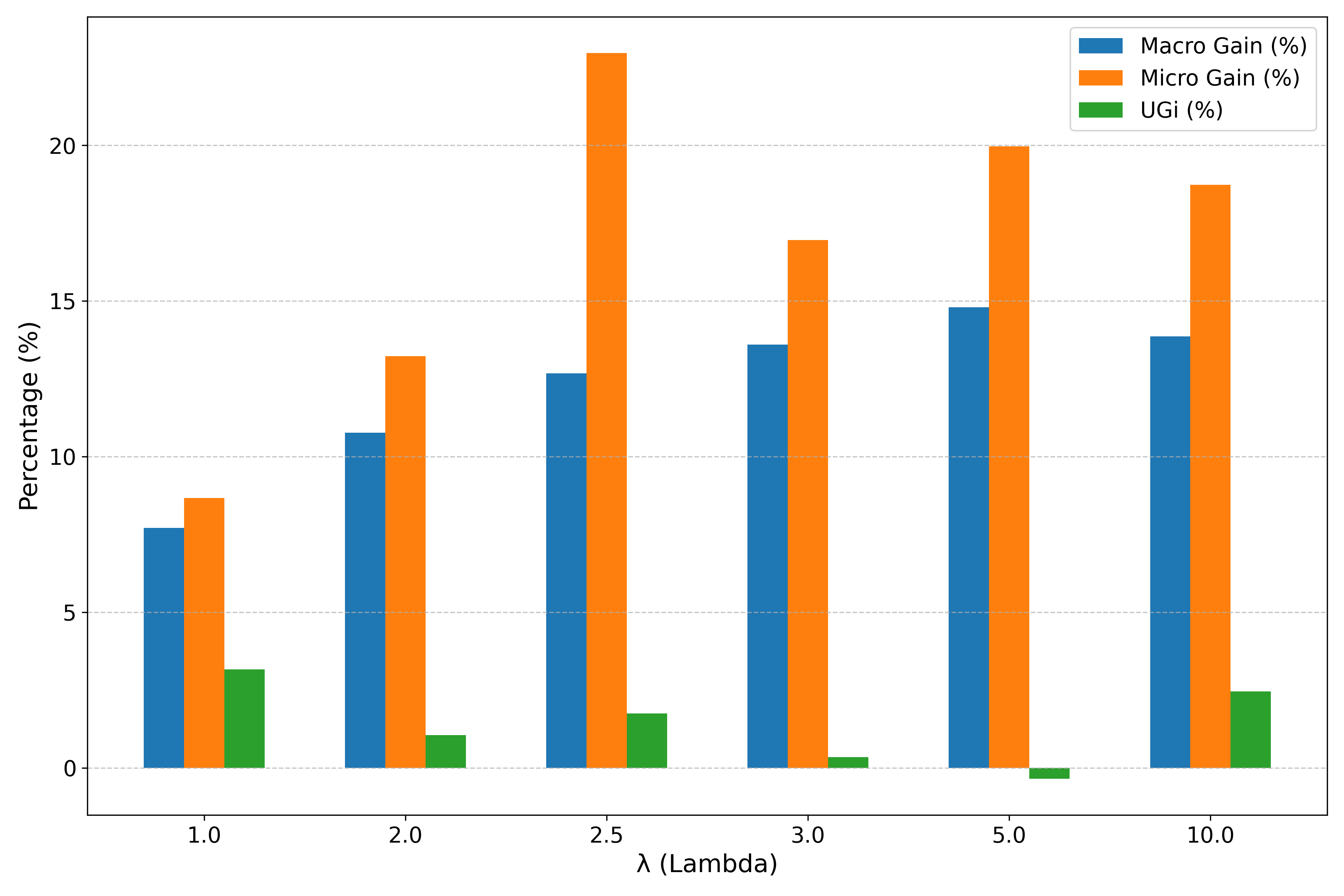} 
    \caption{Comparison of Macro and Micro Gains for Country Across Different Fairness Configurations.}
    \label{fig:countryplot}
    \vskip -10pt
\end{figure}
\captionsetup{font={footnotesize,rm},justification=centering,labelsep=period}%

\begin{figure}[ht]
    \captionsetup{font={footnotesize,rm},justification=centering,labelsep=period}%

    \centering
    \includegraphics[width=0.9\linewidth]{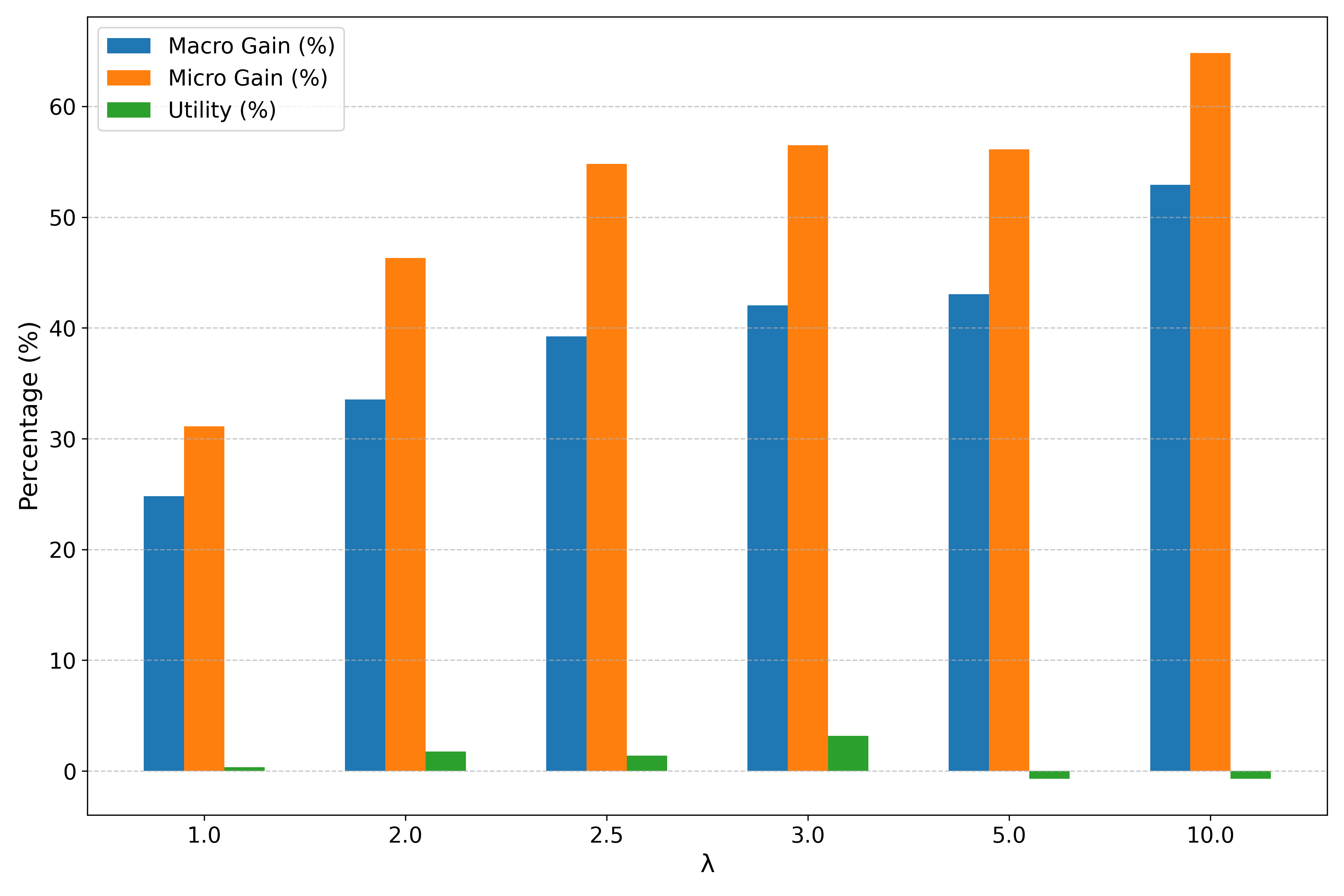} 
    \caption{Comparison of Macro and Micro Gains for Race Across Different Fairness Configurations.}
    \label{fig:raceplot}
    \vskip -10pt
\end{figure}
\captionsetup{font={footnotesize,rm},justification=centering,labelsep=period}%

\subsubsection{Baseline}
We compare our model against a baseline demographic-blind model, which is a conventional (MLP) model that prioritizes quality and ignores fairness constraints. This model selects the original list of papers chosen by the SIGCHI 2017 program committee. By contrasting it with Fair-PaperRec, we identify $\lambda$ values where fairness gains are achieved without utility loss (addressing RQ1). In original settings, the baseline represents the actual peer-review outcomes; improvements over it demonstrate the practical value of fairness interventions in real conferences (addressing RQ2 and RQ3).

\subsubsection{Parameters}
A hyperparameter \(\lambda\) is used to control the trade-off between prediction accuracy and fairness. Higher values emphasize fairness more strongly.
    
The weights \(W_{\text{c}}\) and \(W_{\text{r}}\), respectively, denote the weighting factors assigned to the country and race attributes in the fairness loss function, as shown in Equation~(\ref{eq:combinedloss}).

\begin{figure*}[ht]
    \captionsetup{font={footnotesize,rm},justification=centering,labelsep=period}
    \centering

    \begin{minipage}{0.32\textwidth}
        \centering
        \includegraphics[width=\linewidth]{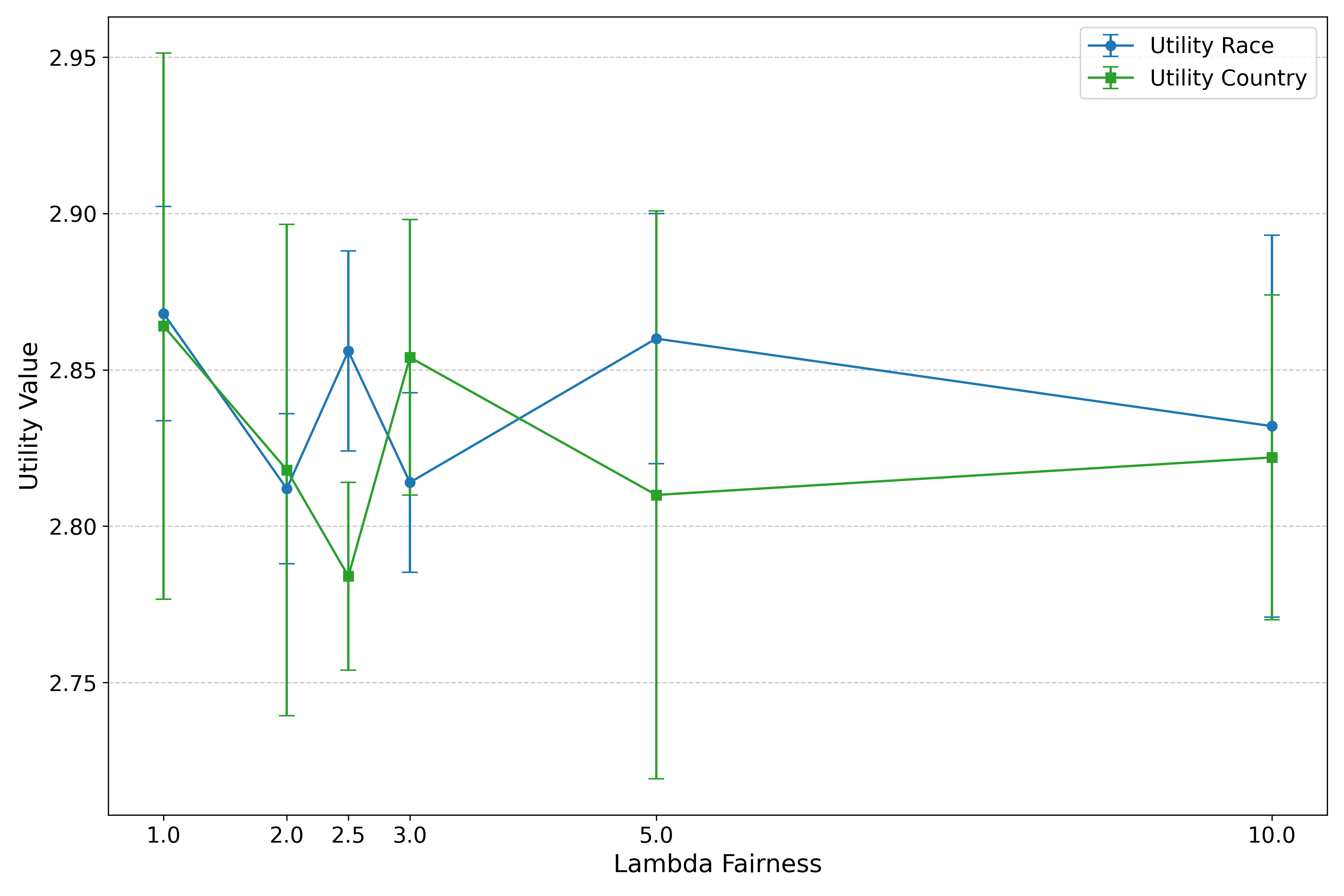}
        \caption*{(a) Utility Gain.}
    \end{minipage}
    \hfill
    \begin{minipage}{0.32\textwidth}
        \centering
        \includegraphics[width=\linewidth]{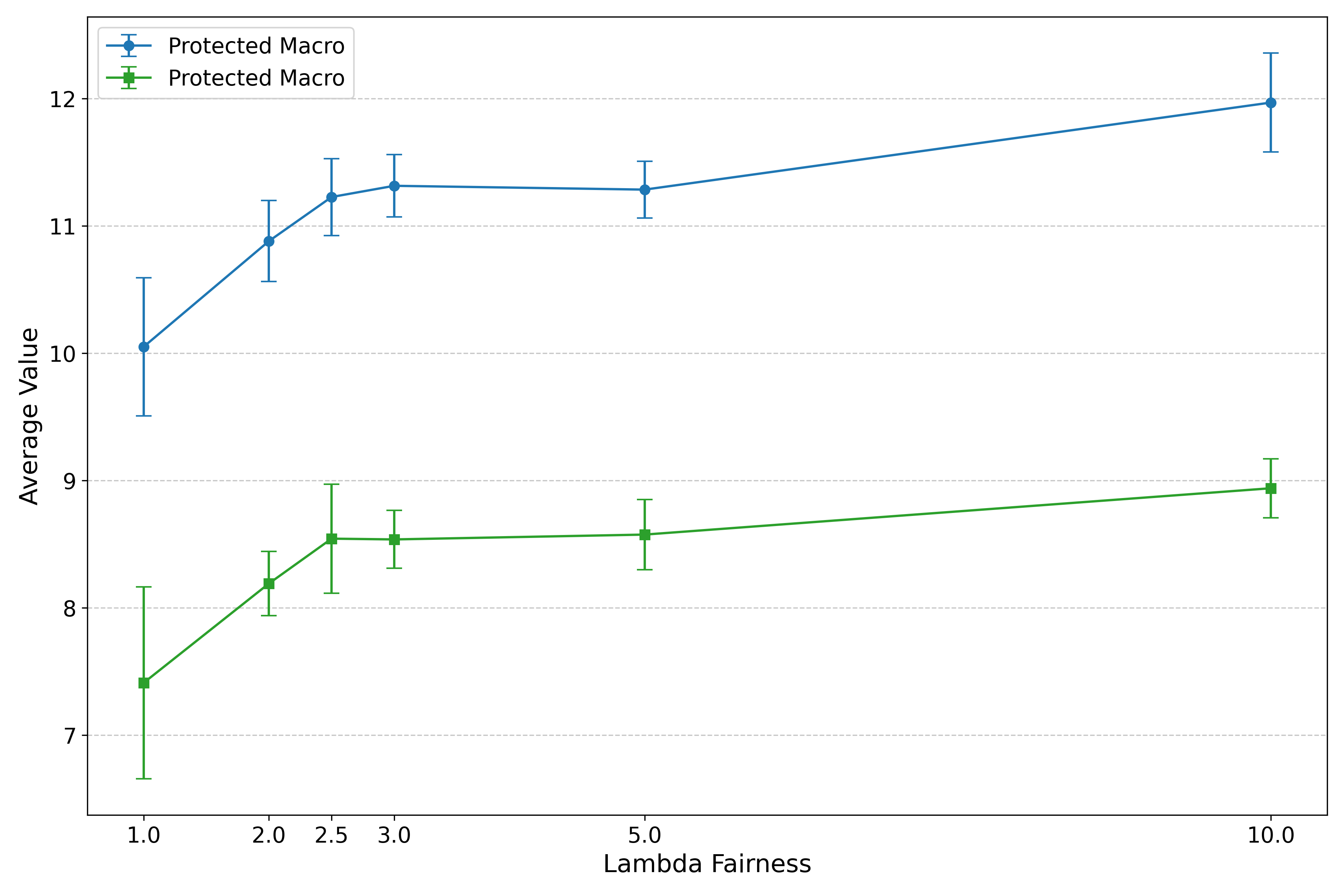}
        \caption*{(b) Macro/Micro for Race.}
    \end{minipage}
    \hfill
    \begin{minipage}{0.32\textwidth}
        \centering
        \includegraphics[width=\linewidth]{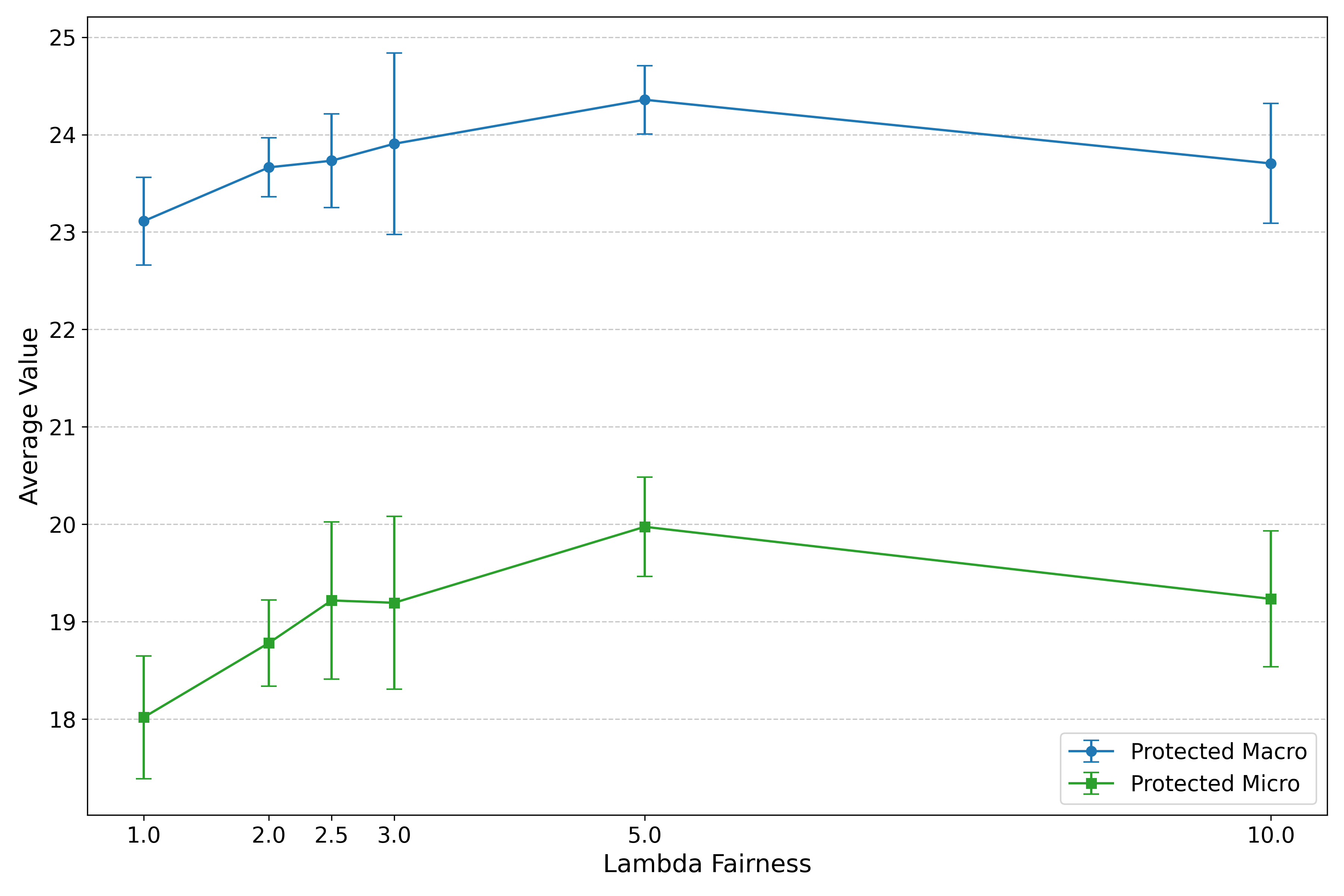}
        \caption*{(c) Macro/Micro for Country.}
    \end{minipage}

    \caption{Comparison of gains across different fairness configurations.}
    \label{fig:combinedsd}
    \vskip -10pt
\end{figure*}
\captionsetup{font={footnotesize,rm},justification=centering,labelsep=period}%

\subsection{Evaluation Metrics}
Diversity is assessed at both the \emph{paper level} and the \emph{author level}. In particular:
\begin{itemize}
\item \emph{Macro Gain} represents the percentage increase in the diversity of each feature within the selected papers compared with the baseline, assessing the overall representation of protected groups.
\item \emph{Micro Gain} is the percentage increase in the diversity of each feature among authors of the selected papers, providing a more detailed perspective on inclusivity.
\item A \emph{Diversity Gain} \cite{alsaffar2021multidimensional} further normalizes these macro-level changes in Equation~(\ref{eq:diversitygain}), capping each feature at 100 to avoid any single attribute skewing the total. 
\item The \emph{F - measure} \cite{alsaffar2021multidimensional} Equation~(\ref{eq:fmeasure}) then combines this diversity improvement with the resulting utility, offering a harmonic balance between fairness gains and paper quality.

\end{itemize}

Table~\ref{table:ablationstudy} summarizes the fairness, diversity, and utility metrics used in our evaluation, including Macro Gain, Micro Gain, Diversity Gain, and the F-measure.

To ensure that diversity enhancements do not compromise scholarly quality, we assess \emph{Utility Gain} ($UG_i$) using a weighted h-index that reflects an author's career stage (professor, associate professor, lecturer, postdoctoral researcher, or graduate student). While the h-index offers a convenient proxy for scholarly influence, it also encodes seniority and field-specific publication practices that may introduce their own biases. For this reason, we interpret utility gain cautiously, not as an absolute measure of “quality,” but as a stability indicator demonstrating whether fairness interventions disproportionately select low impact papers. By comparing the resulting utility values against the baseline distribution, we evaluate whether fairness-aware re-ranking maintains, improves, or undermines the overall quality profile of accepted papers.

\begin{equation}
    D_{G} = \frac{\sum_{i=1}^{n} \min(100, \text{Macro Gain}_{G_{i}})}{n}
    \label{eq:diversitygain}
\end{equation}

\begin{equation}
    F = 2 \times \frac{D_{G} \times (100-UG_{i})}{D_{G} + (100-UG_{i})}
    \label{eq:fmeasure}
\end{equation}

\subsection{Interpretation of the Results}
We evaluate the fairness regularization parameter ($\lambda$) across values 1–10 to examine its impact on fairness (macro and micro gains) and quality (Utility Gain, UGi). The results span both synthetic datasets and original conference datasets. This two-stage design allows us to first test the hypothesis under controlled conditions and then validate it using real-world peer review data.

\paragraph{RQ1: Effect of fairness on utility}
Fig.~\ref{fig:gain_grid} (synthetic) and Figs.~\ref{fig:raceplot}–\ref{fig:countryplot} (original) show that increasing $\lambda$ generally improves macro and micro diversity, while the trajectory of utility differs across settings. Detailed gain calculations are reported in Table~\ref{table:relative_gain_side}.

\begin{itemize}
    \item In synthetic \emph{high-bias} settings, we observe a “sweet spot” ($\lambda \approx 3$) where both fairness and utility peak together. This supports our hypothesis that correcting bias can uncover under-selected, high-quality work.
    \item In synthetic \emph{moderate/fair} settings, small $\lambda$ values act as a useful regularizer, but large values $\lambda$ over-correct, harming utility.
    \item In original data, utility is more stable for race (flat or mildly positive trend) but fluctuates for country, with error bars indicating greater uncertainty at higher $\lambda$. This confirms the trade-off: fairness gains are achievable, but utility must be monitored carefully.
\end{itemize}

\begin{table}[h!]
\centering
\captionsetup{font={footnotesize,rm},justification=centering,labelsep=period}%
\caption{\MakeUppercase{Gain Calculations for Country and Race Features with Utility Gain ($UG_{\text{i}}$).}}
\renewcommand{\arraystretch}{1} 
\setlength{\tabcolsep}{2pt} 
\scriptsize 
\label{table:gaincalc}
\begin{tabular}{l c c c c c c}
    \toprule
    \multicolumn{1}{l}{} & \multicolumn{3}{c}{\textbf{Country Feature}} & 
    \multicolumn{3}{c}{\textbf{Race Feature}} \\
    \cmidrule(lr){2-4} \cmidrule(lr){5-7}
    \textbf{$\lambda$} & \makecell{\textbf{Macro} \\ \textbf{Gain (\%)}} & \makecell{\textbf{Micro} \\ \textbf{Gain (\%)}} & \makecell{\textbf{$UG_{\text{i}}$} \\ \textbf{(\%)}} & \makecell{\textbf{Macro} \\ \textbf{Gain (\%)}} & \makecell{\textbf{Micro} \\ \textbf{Gain (\%)}} & \makecell{\textbf{$UG_{\text{i}}$} \\ \textbf{(\%)}} \\
    \midrule
    1   & 7.71  & 8.67  & 3.16  & 24.81  & 31.11  & 0.35 \\
    2   & 10.77 & 13.23 & 1.05  & 33.54  & 46.30  & 1.75 \\
    2.5 & \textbf{12.67} & \textbf{22.96} & \textbf{1.75}  & 39.25  & 54.81  & 1.40 \\
    3   & 13.60 & 16.96 & 0.35  & \textbf{42.03}  & \textbf{56.48}  & \textbf{3.16} \\
    5   & 14.80 & 19.97 & -0.35 & 43.04  & 56.11  & -0.70 \\
    10  & 13.86 & 18.73 & 2.46  & 52.91  & 64.81  & -0.70 \\
    \bottomrule
\end{tabular}
\vskip -10pt
\end{table}
\captionsetup{font={footnotesize,rm},justification=centering,labelsep=period}%

\paragraph{RQ2: Attribute-specific effects (race vs.\ country)}
The optimal $\lambda$ differs between attributes. Race requires larger $\lambda$ values (around 3) to balance fairness and utility, reflecting the higher initial disparity ratios of racial minorities. Country achieves balance at lower $\lambda$ values (around 2.5). Synthetic experiments revealed the same pattern: the more skewed the initial distribution, the stronger the regularization required. This validates our hypothesis that fairness policies must be tailored to group-specific disparities. A summary of these trade-offs for different $\lambda$ values and weight assignments is provided in Table~\ref{table:ablationstudy}.

\paragraph{RQ3: Multi-attribute and weighting trade-offs}
Experiments that combine race and country with varying weights show that emphasizing one attribute can reduce gains for the other. On synthetic data, joint optimization occasionally diluted improvements compared to single-attribute runs. On original data, this was visible in Micro diversity for country, which exhibited more volatility than race under joint fairness. As summarized in Fig.~\ref{fig:combinedsd}, these configurations reveal how different fairness weightings reshape the balance between utility and diversity, making the trade-offs between race-focused, country-focused, and multi-attribute fairness more explicit.

\paragraph{Macro vs.\ Micro Dynamics}
Across both synthetic and original settings, macro diversity measures were consistently more responsive to fairness regularization than micro measures. Macro gains rose steadily with $\lambda$, especially for race, whereas micro gains were more variable (particularly for country). This suggests that fairness interventions are more effective at improving overall group-level representation than at equalizing outcomes for individual authors.

\paragraph{Conference-level distribution}

The detailed gain calculations for each $\lambda$ are shown in Table~\ref{table:gaincalc}, which reports the breakdown of accepted papers across SIGCHI, DIS, and IUI under different fairness constraints. SIGCHI dominates acceptance at approximately 92\% regardless of $\lambda$, reflecting its strong baseline prestige. However, fairness constraints modestly increase the share of DIS and IUI papers, broadening representation without destabilizing the overall distribution. This indicates that fairness-aware re-ranking can diversify the conference portfolio while preserving dominant trends. For context, the original conference-level distribution of accepted papers is summarized in Table~\ref{table:conf_dist}. In the synthetic stage, the baseline distribution is configured to mimic this original selection profile under controlled bias levels; the corresponding breakdown is reported in Table~\ref{table:synth}.

\subsection{Robustness \& Sensitivity}
For completeness, we report seed variability (mean$\pm$std over multiple random seeds) and, when space allows, confidence intervals around the $\lambda$sweeps to demonstrate that peaks (e.g., Race--High at $\lambda\!\approx\!3$) are stable rather than artifacts of stochasticity.

\begin{table*}[h!]
    \centering
    \captionsetup{font={footnotesize,rm},justification=centering,labelsep=period}%
    \caption{\MakeUppercase{Gain Calculations for Country and Race Features with Utility Gain.}}
    \label{table:ablationstudy}
    \renewcommand{\arraystretch}{1} 
    \setlength{\tabcolsep}{10pt} 
    \resizebox{\textwidth}{!}{%
    \begin{tabular}{c c c c c c c c c}
        \toprule
        \multirow{2}{*}{$\lambda$} &
        \multirow{2}{*}{Weights} &
        \multicolumn{2}{c}{\textbf{Country Feature}} &
        \multicolumn{2}{c}{\textbf{Race Feature}} &
        \multirow{2}{*}{$UG_i$ (\%)} &
        \multirow{2}{*}{Avg.\ $D_G$ (\%)} &
        \multirow{2}{*}{Avg.\ $F$ (\%)} \\

        \cmidrule(lr){3-4} \cmidrule(lr){5-6}
        & & Macro Gain (\%) & Micro Gain (\%) & Macro Gain (\%) & Micro Gain (\%) & & & \\
        \midrule

          & $(W_{\text{r}} = 0.32,\; W_{\text{c}} = 0.68)$ & 6.17  & 6.34  & 30.51 & 46.30 & 3.16 & 44.66 & 53.71 \\
        1 & $(W_{\text{r}} = 1,\; W_{\text{c}} = 2)$       & 6.73  & 9.15  & -0.25 & 0.37  & 2.81 & 6.48  & 13.77 \\
          & $(W_{\text{r}} = 2,\; W_{\text{c}} = 1)$       & 7.43  & 11.43 & 12.91 & 16.11 & 3.16 & 25.63 & 40.36 \\
        \midrule

          & $(W_{\text{r}} = 0.32,\; W_{\text{c}} = 0.68)$ & \textbf{13.60} & \textbf{24.43} & 30.51 & 42.22 & \textbf{4.21} & 55.38 & 68.47 \\
        2 & $(W_{\text{r}} = 1,\; W_{\text{c}} = 2)$       & 5.24  & 6.88  & 15.45 & 17.96 & 0.70 & 20.69 & 21.58 \\
          & $(W_{\text{r}} = 2,\; W_{\text{c}} = 1)$       & 8.36  & 12.86 & 39.49 & 54.26 & 1.75 & 26.31 & 21.58 \\
        \midrule

          & $(W_{\text{r}} = 0.32,\; W_{\text{c}} = 0.68)$ & 8.63  & 17.33 & 36.58 & 50.37 & 2.46 & 56.46 & 66.31 \\
        2.5 & $(W_{\text{r}} = 1,\; W_{\text{c}} = 2)$      & 9.89  & 14.00 & 30.63 & 46.30 & 2.81 & 40.52 & 62.09 \\
          & $(W_{\text{r}} = 2,\; W_{\text{c}} = 1)$       & 9.60  & 17.11 & 42.53 & 56.48 & 1.40 & 59.25 & 69.98 \\
        \midrule

          & $(W_{\text{r}} = 0.32,\; W_{\text{c}} = 0.68)$ & 7.15  & 11.42 & 39.49 & 53.89 & 1.40 & 55.98 & 63.45 \\
        3 & $(W_{\text{r}} = 1,\; W_{\text{c}} = 2)$        & 10.16 & 21.17 & 33.29 & 43.89 & 0.70 & 43.45 & 47.63 \\
          & $(W_{\text{r}} = 2,\; W_{\text{c}} = 1)$        & 9.60  & 18.35 & 42.53 & 55.37 & 2.81 & 61.90 & 47.63 \\
        \midrule

          & $(W_{\text{r}} = 0.32,\; W_{\text{c}} = 0.68)$ & 10.80 & 19.38 & \textbf{45.82} & \textbf{58.52} & 0.70 & \textbf{65.09} & \textbf{72.92} \\
        5 & $(W_{\text{r}} = 1,\; W_{\text{c}} = 2)$        & 4.69  & 3.88  & 33.92 & 40.19 & 0.35 & 38.61 & 15.73 \\
          & $(W_{\text{r}} = 2,\; W_{\text{c}} = 1)$        & 7.43  & 11.90 & 39.49 & 52.96 & 5.26 & 52.26 & 15.73 \\
        \midrule

          & $(W_{\text{r}} = 0.32,\; W_{\text{c}} = 0.68)$ & 9.60  & 18.34 & 42.53 & 55.37 & 1.40 & 62.92 & 70.89 \\
        10 & $(W_{\text{r}} = 1,\; W_{\text{c}} = 2)$       & 7.43  & 13.91 & 24.94 & 25.19 & 4.91 & 32.37 & 34.88 \\
          & $(W_{\text{r}} = 2,\; W_{\text{c}} = 1)$        & 7.43  & 11.72 & 35.44 & 47.41 & -4.21 & 40.53 & 34.88 \\
        \bottomrule
    \end{tabular}
    }
\end{table*}
\captionsetup{font={footnotesize,rm},justification=centering,labelsep=period}%

\subsection{Ablation Study: Multi-Demographic Fairness}

The objective of our ablation study was to evaluate the model's performance when optimizing fairness across multiple demographic attributes simultaneously, specifically with respect to both \emph{country} and \emph{race}. This ablation was conducted to address \emph{RQ3}, which explores the impact of varying fairness weights for each attribute when multiple fairness attributes are considered together.

To ensure fairness, we removed these attributes from the input space, preventing the model from learning direct associations between them and the paper acceptance decisions. Instead, demographic parity loss was computed for each attribute during training, capturing deviations from fairness. The parity losses for both country and race were combined by assigning weights: \(W_c\) for country and \(W_r\) for race, with the initial weights set to \(W_c = 0.68\) and \(W_r = 0.32\), reflecting the distribution of protected groups.

To further explore the model's behavior and answer \emph{RQ3}, we varied these weights, first increasing \(W_c\) while keeping \(W_r\) constant and then increasing \(W_r\) while keeping \(W_c\) fixed. Additionally, we experimented with different values of the fairness regularization parameter \(\lambda\), which controls the trade-off between fairness and utility. These experiments allowed us to observe how different weight configurations and fairness constraints influenced the model’s ability to achieve demographic fairness while maintaining utility and the quality of selected papers.

The results of the ablation study, shown in Table~\ref{table:ablationstudy}, reveal that at \(\lambda = 1\), assigning equal weights to both race and country (\(W_r = 0.32, W_c = 0.68\)) produced significant gains for race, with a Macro Gain of 30.51\% and a Micro Gain of 46.3\%, while country showed relatively smaller improvements (6.17\% and 6.34\%, respectively). However, when the weight for country was increased (\(W_c = 2 \times 0.68\)), diversity gains for race dropped sharply, with a negative macro gain (-0.25\%), while country experienced slight improvements. Conversely, increasing the weight for race (\(W_r = 2 \times 0.32\)) resulted in improved diversity for both race and country, indicating that assigning more weight to race enhances diversity for both attributes to some degree.

At \(\lambda = 2.5\), the model achieved the best balance between diversity and utility. Equal weights for race and country yielded macro and micro gains of 36.58\% and 50.37\% for race, and 8.63\% and 17.33\% for country, with a low utility loss of 2.46\%. This suggests that \(\lambda = 2.5\) is optimal for balancing fairness and utility. As \(\lambda\) increases further, race diversity continues to improve (reaching 45.82\% Macro Gain at \(\lambda = 5\)), but at the cost of decreasing utility. The different optimal \(\lambda\) values for race and country suggest that disparity ratios impact how fairness constraints should be weighted, with race requiring a higher \(\lambda\) due to its higher disparity ratio. This leads to greater race diversity gains at higher \(\lambda\) values, whereas country achieves optimal results at moderate \(\lambda\) values, such as 2.5.

These findings directly address \emph{RQ3}, demonstrating that fairness weights must be carefully calibrated for each protected attribute. Assigning greater weight to race tends to improve diversity for both race and country, whereas increasing the weight for country may result in reduced fairness for race. The optimal balance between fairness and utility is achieved when fairness weights and \(\lambda\) values are adjusted based on the unique disparity ratios of each attribute.

\section{Conclusion and Future Work}

We began our experiment with a hypothesis tested on synthetic data: a lightweight fairness regularizer, controlled by $\lambda$, could improve demographic inclusion without compromising quality. Applying the same regularization to the original conference data confirmed and strengthened this insight: fairness regularization consistently improved representation, particularly for race, while leaving utility largely stable or mildly positive.

These results highlight three key lessons. First, fairness effects depend on the degree of underlying disparity: high-bias settings benefit from stronger regularization, while near-fair regimes require small adjustments to avoid over-correction. Second, attributes vary in sensitivity: race often demands more intervention than country, and multi-attribute optimization requires careful balancing. Third, in strongly biased systems, fairness and quality are not in conflict; debiasing uncovers under-selected, high-quality work.

At the same time, our approach has limits. We rely on a simple MLP with post-hoc fairness loss, excluding protected attributes from the input but not explicitly modeling causal pathways or reviewer dynamics. Extending this framework with causal inference, graph-based architectures capturing author–institution–topic relations, or generative approaches such as VAEs could deepen bias mitigation while preserving interpretability. Broader definitions of fairness, adaptive tuning of $\lambda$ and attribute weights, and human-in-the-loop evaluation are promising directions. Embedding fairness interventions into real-world workflows will require robustness to distribution shifts, transparency in reporting, and governance mechanisms that account for privacy and accountability. Together, these steps can transform Fair-PaperRec from a research prototype into a practical, equitable tool for the future of scholarly peer review.

\section*{Acknowledgment}
This work was supported by the National Science Foundation (NSF) under Award number OIA-1946391, Data Analytics that are Robust and Trusted (DART).

\printbibliography
\end{document}